\documentclass{article}
\pdfoutput=1
\usepackage[preprint]{corl_2023}
\usepackage[utf8]{inputenc} % allow utf-8 input
\usepackage[T1]{fontenc}    % use 8-bit T1 fonts
\usepackage{hyperref}       % hyperlinks
\usepackage{url}            % simple URL typesetting
\usepackage{booktabs}       % professional-quality tables
\usepackage{amsfonts}       % blackboard math symbols
\usepackage{nicefrac}       % compact symbols for 1/2, etc.
\usepackage{microtype}      % microtypography
\usepackage{lipsum}
\usepackage{fancyhdr}       % header
\usepackage{graphicx}       % graphics
\usepackage{amsmath}
\usepackage{amsmath,amssymb,amsfonts}
\usepackage{subcaption}\usepackage{times}  % DO NOT CHANGE THIS
\usepackage{helvet}  % DO NOT CHANGE THIS
\usepackage{courier}  % DO NOT CHANGE THIS
\usepackage{multirow}
\usepackage{multicol}
\usepackage{wrapfig}
\usepackage{diagbox}
\usepackage[table,xcdraw]{xcolor}
\usepackage{array}
\usepackage{natbib}  % DO NOT CHANGE THIS AND DO NOT ADD ANY OPTIONS TO IT
\usepackage{caption} % DO NOT CHANGE THIS AND DO NOT ADD ANY OPTIONS TO IT
\graphicspath{{media/}}     % organize your images and other figures under media/ folder

\title{Vehicle Dynamics Modeling for Autonomous Racing Using Gaussian Processes
}

\author{
  Jingyun Ning \\
  Dept. of Electrical and Computer Engineering \\
  Univ. of Virginia \\
  \texttt{jn2ne@virginia.edu} \\
   \And
  Madhur Behl \\
  Dept. of Computer Science \\
  Univ. of Virginia \\
  \texttt{madhur.behl@virginia.edu} \\
}

\begin{document}
\maketitle

\begin{abstract}
Autonomous racing is increasingly becoming a proving ground for autonomous vehicle technology at the limits of its current capabilities. The most prominent examples include the F1Tenth racing series, Formula Student Driverless (FSD), Roborace, and the Indy Autonomous Challenge (IAC). Especially necessary, in high speed autonomous racing, is the knowledge of accurate racecar vehicle dynamics. The choice of the vehicle dynamics model has to be made by balancing the increasing computational demands in contrast to improved accuracy of more complex models.
Recent studies have explored learning-based methods, such as Gaussian Process (GP) regression for approximating the vehicle dynamics model. 
However, these efforts focus on higher level constructs such as motion planning, or predictive control and lack both in realism and rigor of the GP modeling process, which is often over-simplified. 
This paper presents the most detailed analysis of the applicability of GP models for approximating vehicle dynamics for autonomous racing.  In particular we construct dynamic, and extended kinematic models for the popular F1TENTH racing platform. 
We investigate the effect of kernel choices, sample sizes, racetrack layout, racing lines, and velocity profiles on the efficacy and generalizability of the learned dynamics. We conduct 400+ simulations on real F1 track layouts to provide comprehensive recommendations to the research community for training accurate GP regression for single-track vehicle dynamics of a racecar. 
\end{abstract}

\section{Introduction}
%In the past decade, autonomous vehicle racing has gained considerable attention due to the advances in computing and sensing technologies.
The rising popularity of self-driving cars has led to the emergence of a new research field in recent years: Autonomous racing. Researchers are developing algorithms for high-performance race vehicles which aim to operate autonomously on the edge of the vehicle's limits: High speeds, high accelerations, low reaction times, highly uncertain, dynamic, and adversarial environments. Competitions in autonomous racing have been held not only in simulators \cite{hartmann2021autonomous, babu2020f1tenth}, but also on hardware with racecars' ranging from 1:43 scale RC cars~\cite{carrau2016efficient} to 1/10 scale (F1tenth Racing~\cite{o2019f1}) to full size Indy racecars \cite{ wischnewski2022indy}. 

The modeling of the vehicle dynamics behavior of the racecar is an essential part in the field of autonomous racing.
The current state of the art provides many variations of vehicle dynamics modeling such as single track model, double track model or full vehicle model. The more complicated the vehicle dynamics model, the more parameters are needed. Unfortunately not all of those parameters are available in detail for a vehicle and so different methods for estimating these parameters are proposed - especially for nonlinear vehicle parameters like the tires.
A major challenge is the model mismatch between mathematical model and real vehicle dynamics. While researches in vehicle modeling have developed several models, from simplest point-mass model to fully multi-body model \cite{althoff2016set, tomas2013vehicular}, the modeling errors are inevitable due to highly non-linearity of real vehicle dynamics.

Consequentially, implementation of learning-based methods to compensate the error between simulation models and real vehicle models has received attention from researchers. Gaussian Processes (GP) based models are powerful candidates among such methods. Previous work \cite{jain2020bayesrace} on autonomous racing has explored implementing GP regression for vehicle dynamics modeling. 
However, their exploration is limited and lacks rigorous analysis - such as knowledge of GP model kernel selection, sample size, different racing lines and track layouts. 
During GP model selection, both the form of the mean function and covariance kernel function need to be carefully chosen and tuned. 
While the mean function is typically constant, either zero or the mean of the training dataset. 
There are many options for the covariance kernel functions: they can have many forms as long as it follows the properties of a kernel. 
Moreover, autonomous racing specific criteria such as vehicle velocity profiles and racetrack curviness are often omitted when implementing GP for vehicle modeling. 

%For example, previous work tends to use synthetic tracks, which are made of curvy racetracks which make them suitable for GP prediction of vehicle dynamics.

This paper presents the most detailed analysis of the suitability of GPs for vehicle dynamics modeling for autonomous racing to date.
We build a dynamic and extended kinematic (E-Kin) model of the 1/10 scale racecar using a realistic simulation platform (F1TENTH Gym). 
In addition, we provide a comprehensive study for GP model selection based on:  
\begin{enumerate}
    \item Vehicle dynamics model in three real-world Formula One racetracks: Shanghai International Circuit, Sepang International Circuit, and Yas Marina Circuit. \par
\item Different driving scenarios combinations: Different velocity profiles and racelines during the data collection and their effect on GP accuracy.\par
\item Exploration of suitable combinations of five GP kernel functions.\par
\item Analysis of the effect of training sample size on the model accuracy. \par
\end{enumerate}

To answer the above questions, we provide a comprehensive and rigorous analysis of GPs for vehicle dynamic modeling by conducting 400+ simulations.

% \begin{figure*}[htp]
%      \centering
     
%      \begin{subfigure}[b]{0.25\textwidth}
%          \centering
%          \includegraphics[width=\textwidth]{LaTeX/Images/Shanghai_map.png}
%          \caption{Shanghai International Circuit}
%          \label{fig:SH}
%      \end{subfigure}
%      \hfill
%      \begin{subfigure}[b]{0.25\textwidth}
%          \centering
%          \includegraphics[width=\textwidth]{LaTeX/Images/Sepang_map.png}
%          \caption{Sepang International Circuit}
%          \label{fig:SP}
%      \end{subfigure}
%      \hfill
%      \begin{subfigure}[b]{0.25\textwidth}
%          \centering
%          \includegraphics[width=\textwidth]{LaTeX/Images/YasMarina_map.png}
%          \caption{Yas Marina Circuit}
%          \label{fig:YM}
%      \end{subfigure}
%         \caption{Simple graphs of three real-world F1 racetracks.}
%         \label{fig:tracks}
% \end{figure*}

\section{Background: Vehicle Dynamics Modeling}
We consider the single-track model in this paper, in which the two front wheels as well as the two rear wheels are lumped into one wheel each. Both kinematic model and dynamic model has been described in this section. Following simplifications have been made: (i) we assume the vehicle to have planar motion, (ii) as the car used is rear wheel driven, the longitudinal force on the front wheel is neglected.

\begin{table*}[t]
\centering
\scriptsize
\begin{equation*}
\renewcommand{\arraystretch}{1.5}
\mit
\begin{tabular}{|c|ccc|}
\hline
\textbf{Notations}                              & \multicolumn{3}{c|}{\textbf{Vehicle Dynamics}}                                                                                                                                                                                        \\ \hline
$x$,$y$: vehicle position w.r.t. inertial frame  & \multicolumn{1}{c|}{Kinematic}                                       & \multicolumn{1}{c|}{Dynamic}                                                              & E-Kin                                                              \\ \hline
$v$: Velocity [$m/s$]                           & \multicolumn{1}{c|}{$\dot x = v \cos(\psi)$}                         & \multicolumn{1}{c|}{$\dot x = v\cos(\psi+\beta)$}                                         & $\dot x = v\cos(\psi+\beta)$                                       \\ \hline
$\delta$: Steering angle [$rad$]                & \multicolumn{1}{c|}{$\dot y = v \sin(\psi)$}                         & \multicolumn{1}{c|}{$\dot y = v\sin(\psi+\beta)$}                                         & $\dot y = v\sin(\psi+\beta)$                                       \\ \hline
$\psi$: Heading angle [$rad$]                   & \multicolumn{1}{c|}{$\dot \delta = v_{\delta}$}                      & \multicolumn{1}{c|}{$\dot \delta = v_{\delta}$}                                           & $\dot \delta = v_{\delta}$                                         \\ \hline
$\omega$: Yaw rate [$rad/s$]                    & \multicolumn{1}{c|}{$\dot v = a_{long}$} & \multicolumn{1}{c|}{$\dot v = a_{long} = \frac{F_{rx}}{m}$}                               & $\dot v = a_{long}$                                                \\ \hline
$\beta$: Body slip angle [$rad$]                & \multicolumn{1}{c|}{$\dot \psi = \frac{v}{l_{r}+l_{f}}\tan(\delta)$}                             & \multicolumn{1}{c|}{$\dot \psi = \omega = \frac{v}{l_{r}+l_{f}}\tan(\delta)$}             & $\dot \psi = \omega$                                               \\ \hline
$a_{long}$: Longitudinal acceleration [$m/s^2$] & \multicolumn{1}{c|}{}                                                & \multicolumn{1}{c|}{$\dot \omega = \frac{1}{I_z}(l_{f}F_{fy}\cos\delta_{f}-l_{r}F_{ry})$} & $\dot \omega = \frac{1}{l_r + l_f}(\dot \delta v + \delta \dot v)$ \\ \hline
$\delta_v$: Steering velocity [$rad/s$]         & \multicolumn{1}{c|}{}                                                & \multicolumn{1}{c|}{$\dot \beta = \frac{1}{mv}(F_{fy}+F_{ry})-\omega$}                    & $\dot \beta = (l_r + l_f)(\dot \delta v + \delta \dot v)$          \\ \hline
\end{tabular}
\end{equation*}
\caption{Mathematical descriptions of different vehicle dynamics models}
\label{equations}
\end{table*}

\subsection{Kinematic single-track model}
Kinematic single-track model is preferred in applications for its simplicity ~\cite{thrun2006stanley, kong2015kinematic}. We adopt the kinematic model from ~\cite{althoff2017commonroad}.

The kinematic model 
% has three parameters, vehicle length $l$, width $w$, and wheelbase length $l_{wb}$. It 
requires two tuning parameters, $l_r$ and $l_f$, which represent the distance from center of gravity (C.O.G.) to the rear and front axles, respectively. The differential equations for such a model are given in Table~\ref{equations}. $x$ and $y$ are the location of C.O.G. with respect to the inertial frame. Velocity at the C.O.G. of the vehicle is denoted by $v$. $\Psi$ is vehicle inertial heading orientation, and $a_{long}$ is longitudinal acceleration. $\delta$ is steering angle represents the angle between vehicle heading and front wheel direction. 

\subsection{Single-track model}
\begin{wrapfigure}[16]{r}{0.45\textwidth}
    \centering
    \includegraphics[width=0.45\textwidth]{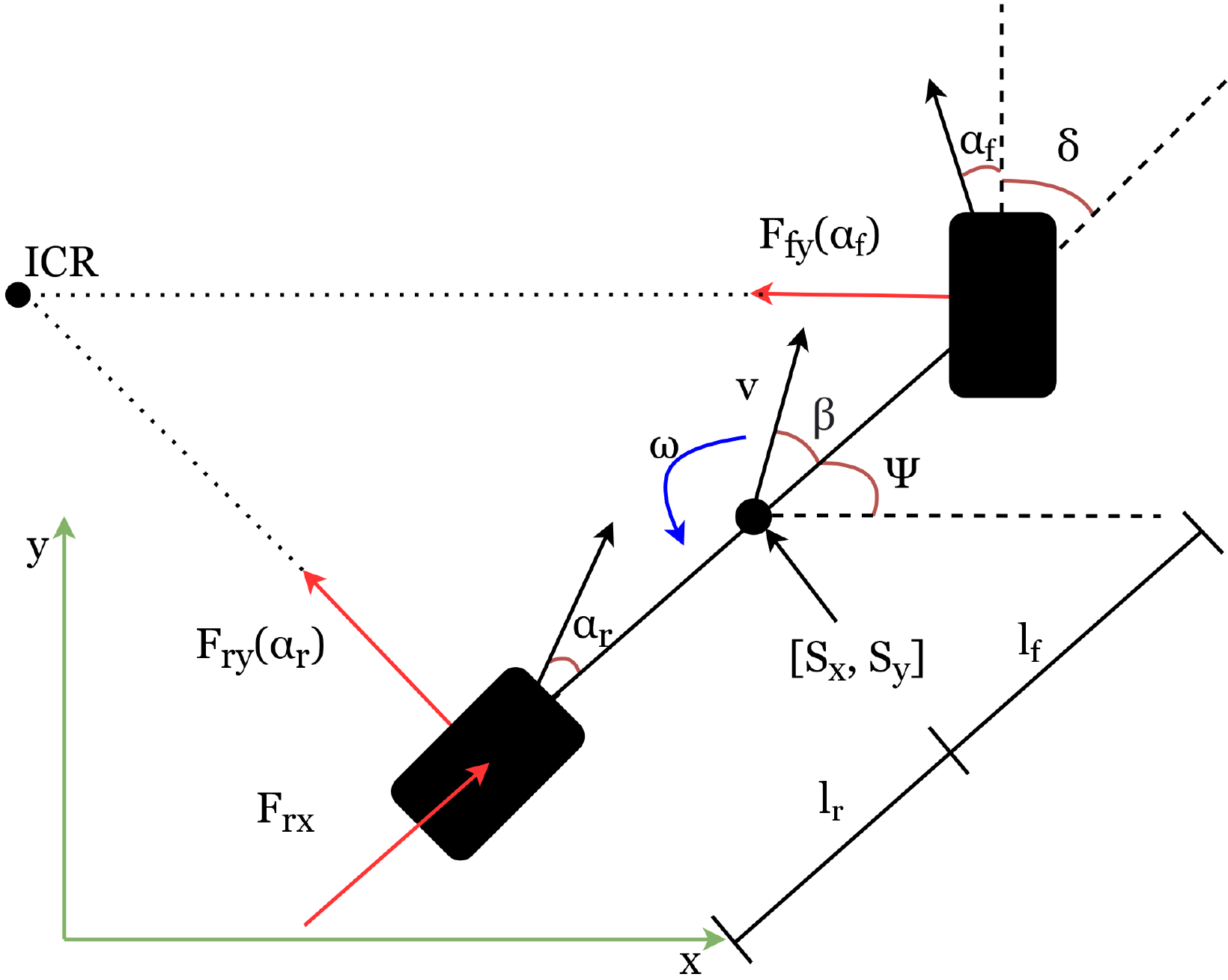}
    \caption{Single-track dynamic vehicle model. Reference point: C.O.G.}
    \label{fig:dyn}
\end{wrapfigure}

At higher speeds, kinematic models can not be applied since they do not consider tire slip, which means important effects such as understeer or oversteer are not considered~\cite{rajamani2011vehicle}. Therefore, when performing planning of evasive maneuvers closer to physical limits, a single-track (dynamic) model needs to be developed for both lateral and longitudinal control~\cite{hwan2011anytime, shiller1991dynamic}, shown in Fig.~\ref{fig:dyn}. The slip angle of a tire is defined as the angle between the orientation of the tire and the orientation of the velocity vector of the wheel. 
% We use $\alpha_r$ and $\alpha_f$ denote rear tire slip angle and front tire slip angle, respectively. 
The mathematical descriptions are denoted in Table~\ref{equations}, where

% \begin{subequations}
% \label{eq:dyn_m}
% \begin{alignat}{2}
% \dot X &= v\cos(\psi+\beta) \label{sub-eq-1:dyn_m}\\
% \dot Y &= v\sin(\psi+\beta) \label{sub-eq-2:dyn_m}\\
% \dot \delta &= v_{\delta} \label{sub-eq-3:dyn_m}\\
% \dot v &= a_{long} = \frac{F_{rx}}{m}\label{sub-eq-4:dyn_m}\\
% \dot \psi &= \omega = \frac{v}{l_{r}+l_{f}}\tan(\delta) \label{sub-eq-5:dyn_m}\\
% \dot \omega &= \frac{1}{I_z}(l_{f}F_{fy}\cos\delta_{f}-l_{r}F_{ry}) \label{sub-eq-6:dyn_m}\\
% \dot \beta &= \frac{1}{mv}(F_{fy}+F_{ry})-\omega \label{sub-eq-7:dyn_m}
% \end{alignat}
% \end{subequations}

% The derivatives \eqref{sub-eq-1:dyn_m}-- \eqref{sub-eq-5:dyn_m} are obtained as for the kinematic model, \eqref{sub-eq-6:dyn_m} is obtained by computing the derivative of \eqref{sub-eq-5:dyn_m} of the kinematic model, and the \eqref{sub-eq-7:dyn_m} is determined by differentiating the body slip angle $\beta$. 

\begin{subequations}
\small
\label{eq:dyn_f}
\begin{alignat}{2}
F_{yf} &= 2\mu C_{sf}\frac{(mgl_r - ma_{long}h_{cog})}{l_f + l_r}\alpha_{f} \label{sub-eq-1:dyn_f}\\
F_{yr} &= 2\mu C_{sr}\frac{(mgl_f + ma_{long}h_{cog})}{l_f + l_r}\alpha_{r} \label{sub-eq-2:dyn_f}    
\end{alignat}
\end{subequations}

\begin{subequations}
\label{eq:dyn_a}
\begin{alignat}{2}
\alpha_{f} &= \delta - \beta - \frac{l_{f}\omega}{v} \label{sub-eq-1:dyn_a}\\
\alpha_{r} &= (- \beta + \frac{l_{r}\omega}{v}) \label{sub-eq-2:dyn_a}    
\end{alignat}
\end{subequations}

$F_{rx}$ is the longitudinal force at the rear axle, and $F_{fy}$, $F_{ry}$ are the lateral forces at front axle and rear axle, respectively. Their governing equations are described as \eqref{sub-eq-1:dyn_f} and \eqref{sub-eq-2:dyn_f}, where $\alpha_{f}$ and $\alpha_{r}$ are tire slip angles at front wheels and rear wheels, respectively.

\subsection{Extended kinematic model}
A well-tuned single-track model is suitable for autonomous racing with advanced control algorithms. However, due to the model complexity, the model tuning procedure is time prohibitive. In addition, we need access to the tire models for calculating lateral forces, and the tire models need to be recalibrated each time for a new racetrack, which increases the cost of implementing such a dynamic model. While a simpler kinematic model is enough for low-speed driving behaviors, the model mismatch will become significant at much higher speeds. 

Therefore, in this work, we adopt learning-based extended kinematic (E-Kin) model, which has same states as a dynamic model, and we use GP models to calculate the discrepancies between models.
Mathematical representation is shown in Table~\ref{equations}. The differences between E-kin and dynamic model lie in yaw rate, $\omega$, and vehicle body slip angle, $\beta$, as shown in Eq.~\ref{eq:ext_k}.

\begin{subequations}
\label{eq:ext_k}
\begin{alignat}{2}
\dot \omega &= \frac{1}{l_r + l_f}(\dot \delta v + \delta \dot v) \label{sub-eq-6:ext_k}\\
\dot \beta &= (l_r + l_f)(\dot \delta v + \delta \dot v) \label{sub-eq-7:ext_k}
\end{alignat}
\end{subequations}

\section{Related Work}

GP regression models have become popular for autonomous driving in the past decade, especially in autonomous racing. They can be implemented to predict ego and target vehicle behaviors in multi-agent scenarios when the ego vehicle performs an overtake maneuver.

The work of Hewing et al makes use of a sparse GP approximation with dynamically adjusting inducing inputs and enable a real-time implementable NMPC controller to increase the racing performance. This method aims to learn from vehicle sensor data with GP to improve the ego vehicle dynamics model \cite{hewing2018cautious}. 

Wischnewski et al. \cite{wischnewski2019model} present the implementation of GP for a nonlinear regression problem. This approach tends to mitigate the gap between planned and driven trajectory which is caused by the control system quality and unmodelled effects.

\cite{busch2022gaussian, brudigam2021gaussian} use GP models to predict future target vehicle behaviors in a head-to-head racing environment. Those work employs the GP models to predict target vehicle trajectories learned from previous behaviors for ego vehicles to decide whether to perform overtaking maneuvers or not.

The extended kinematic model in this paper is inspired by Jain et al. \cite{jain2020bayesrace}. In their work, they utilized the GP regression model to identify uncertainties between the extended kinematic and dynamic models and demonstrated the capability of MPC implementation on their extended kinematic model.

All of the related work have utilized GP models for either predicting target vehicle trajectories or identifying errors of ego vehicle states from a simpler vehicle model. However, they are limited in following aspects:
\begin{enumerate}
    \item The lack of knowledge of GPs kernel function selection;
    \item The impact on different sample sizes for training GPs;
    \item The effect of utilizing different racing lines and speed profiles;
    \item Racetracks are not realistic.
\end{enumerate}

\section{Problem Statement}
\subsection{List of Assumptions}
\subsubsection{Assumption 1: 2D planar}
This paper assumes the vehicle is driving in two-dimensional space, i.e., only yaw angles are being considered. In the future, this assumption can be relaxed by selecting racetracks that have banked curves, which will introduce the pitch angles in addition to yaw angles.

\subsubsection{Assumption 2: Full knowledge of the vehicle model}
We use a single-track dynamic model to represent the real vehicle model and assume that the F1TENTH Gym data of the dynamic model corresponds to the ground truth. This simplification can be relaxed by either implementing a more sophisticated vehicle model, e.g., a multi-body vehicle model, or using a realistic simulator, e.g., an SVL simulator.

\subsubsection{Assumption 3: Perfect sensor measurements}
The racecar in the F1TENTH Gym simulator includes on-board LIDAR and IMU, which can estimate the wheel odometry. This paper assumes the vehicle has sufficient sensors to measure the states of interest for training GPs, and there are no sensor noises associated with data collection. This assumption could be relaxed in the future by introducing random noises to the sensor measurements.

\subsection{Problem formulation}
In this paper, we have not only built a dynamic model and extended kinematic model of the F1TENTH racecar, but also trained GP regression model to learn the mismatch between these models. We build the corrected kinematic model, $f_{corr}$, as an approximation of the dynamic model: 
\begin{equation}
    f_{dyn} \sim f_{corr}
\end{equation}
where the corrected model is a combination of E-kin model, $f_{E-Kin}$, and error model, $e$.
\begin{equation}
f_{corr}(x_k,u_k) = f_{E-Kin}(x_k,u_k) + e(x_k,u_k) 
\end{equation}

% Although GP models have been widely used in previous work, no work has studied GP models in detail for vehicle dynamics modeling. In addition to learn the model mismatch with GP, we also provide an in-depth analysis of GP implementations.

We define the form of error model in Eq.~\ref{eq:error}:

\begin{equation}
\begin{aligned}
   e(x_k, u_k) = x_{k+1} - f_{E-Kin}(x_k, u_k)
\end{aligned}
\label{eq:error}
\end{equation}

$x$ is the vehicle sensor measurements (states), consists of: $[X,Y,v,\omega,\beta]$, $u$ is control inputs, which are  $[a_{long}, \delta_v]$, and $f_{E-Kin}$ is the estimates which are calculated by the E-Kin model. Since the model mismatches are only in the yaw rate $\omega$, and body slip angle, $\beta$, we define the formula of GP learned error model as in~\ref{eq:errorGP}.

\begin{equation}
\begin{aligned}
    e = \mathcal{GP}(\omega, \beta, a, \delta_v)
\end{aligned}
\label{eq:errorGP}
\end{equation}

In the next section, we will describe the methods of generating the data set for GP to learn the error models.

% \subsection{Gaussian Process Model}
% Gaussian process (GP) regression is a powerful non-parametric tool used to infer values of an unknown function given previously collected measurements, and it has been widely used in autonomous racing field. Given a training set with $n$ sample points ${(x_1,y_1), ..., (x_n, y_n)}$, one can use GP to model a non-linear function $y=f(x)
% % :\mathbf{R}^d \mapsto \mathbf{R}
% $. We present following mathematical background based on \cite{seeger2004gaussian}.

% GP can be specified by its mean function $m(x)$ and covariance function $k(x,x')$ of a real process $f(x)$
% \begin{equation}
% \label{gp}
% \begin{aligned}
%     f(x) \sim \mathcal{GP}(m(x), k(x,x'))
% \end{aligned}
% \end{equation}
% where, 
% \begin{equation}
% \label{mk}
% \begin{aligned}
%     m(x) &=\mathrm{E}[f(x)], \\
%     k(x,x') &=\mathrm{E}[(f(x)-m(x))(f(x')-m(x'))],
% \end{aligned}
% \end{equation}
% The model predictions can be drawn by conditioning the prediction point on the observation set. The greatest practical advantage of using GP models is they can give a reliable estimate of the uncertainty of their predictions. This characteristic of GP helps us to explore different control strategies during the optimization process. Another advantage of GP is that it allows one to incorporate expert knowledge. When modeling a problem, one can choose the kernel functions based on their knowledge of the shape of functions output. In this paper, we have studied and evaluated kernel composing for vehicle modeling.\par

\section{Methodologies}

\subsection{Model selection}
The goal of our method is to explore the best setting for training GPs for vehicle dynamic modeling in autonomous racing. In other words, we want to evaluate the effect of different kernel functions, sample sizes, racing scenarios and racetrack layouts on training GP models.

\begin{figure}
    \centering
    \includegraphics[width=0.7\columnwidth]{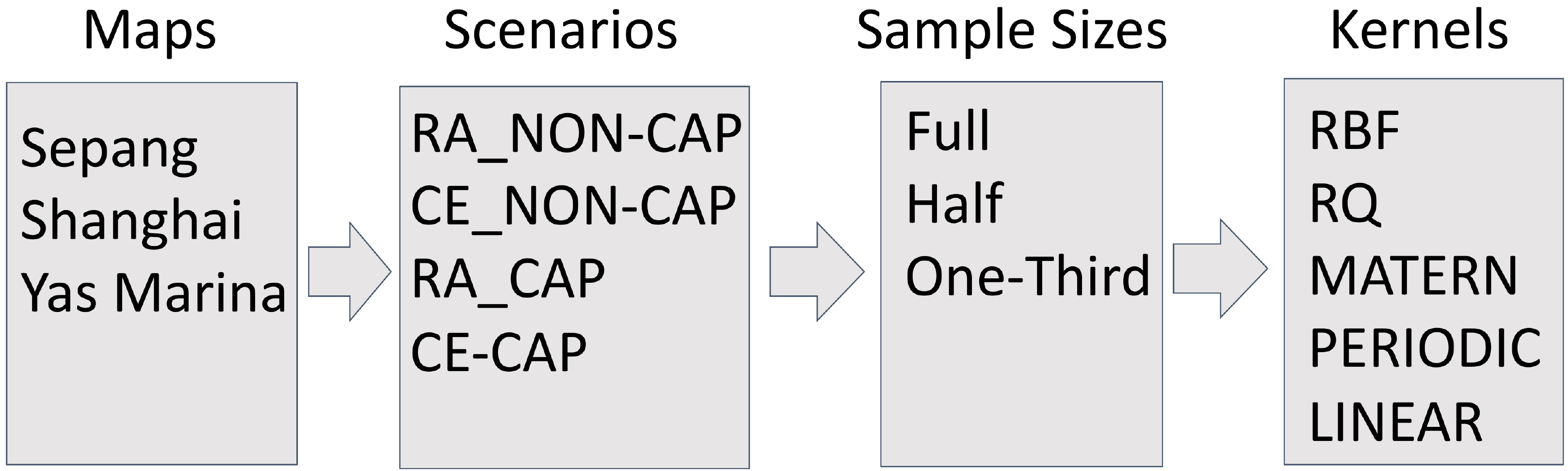}
\caption{Workflow of exploring the best settings for GP training for vehicle dynamic modeling.}
\label{fig:pipe1}
\end{figure}

For the model selection, we design workflow as depicted in Fig~\ref{fig:pipe1}: For each Formula One tracks, we collect training data of four different racing scenarios with different sample sizes. Then we train GP models with different combinations of five standard kernel functions. Thus, to decide the best model setting, we have conducted a thorough $3 \times 4 \times 3 \times 12 = 432$ experiments. The details of model schemes analysis are shown in Table \ref{tab:analysis}.

\subsection{Kernel composing}
Kernel composing is the standard method for training GP models with more than one type of feature. \cite{duvenaud2014automatic}. Since the GP models learned in this paper has different features of input data, acceleration, $a_{long}$ and steering velocity, $\delta_v$. We have selected five standard kernel functions, Table \ref{tab:kernels}, to explore the effect of different kernel composing on training GP models. The multiplication and addition are denoted as following

\begin{table}
\centering
\scriptsize
\renewcommand{\arraystretch}{1.2}
% \resizebox{\columnwidth}{!}{
\begin{tabular}{|c|c|}
\hline
\textbf{Standard Kernel Functions} & \textbf{Mathematical Definition}                                                                                                                      \\ [8pt]\hline 
RBF                       & $k_{RBF}(x, x') = \sigma^2\mathbf{exp}(- \frac {(x-x')^2}{2l^2})$                                                                            \\ [8pt]\hline 
RQ                        & $k_{RQ}(x, x') = \sigma^2(1 + \frac {(x-x')^2}{2\alpha l^2})$                                                                       \\ [8pt]\hline 
PERIODIC                  & $k_{PER}(x, x') = \sigma^2\mathbf{exp}(- \frac {2sin^2(\pi|x-x'|/p)}{l^2})$                                                        \\ [8pt]\hline 
LINEAR                    & $k_{LIN}(x, x') = \sigma_{b}^2 + \sigma_{v}^2(x-c)(x'-c)$                                                                                    \\ [8pt]\hline 
MATERN                    & $k_{MAT}(x,x') = \frac{1}{\Gamma(v)2^{\nu-1}}(\frac{\sqrt{2\nu}}{l}d(x, x'))^\nu K_{\nu}(\frac{\sqrt{2\nu}}{l}d(x, x'))$ \\ [8pt] \hline 
\end{tabular}
% }
\caption{Standard kernel functions and math definition.}
\label{tab:kernels}
\end{table}

% \begin{subequations}\label{eq:kernelcombo}
%     \begin{alignat}{2}
%     k_a + k_b &= k_a(x,x') + k_b(x, x') \label{sub-eq-1:kernelcombo}\\
%     k_a \times k_b &= k_a(x,x') \times k_b(x, x')\label{sub-eq-2:kernelcombo}
%     \end{alignat}
% \end{subequations}

\begin{table}
\renewcommand{\arraystretch}{1.2}
\scriptsize
\centering
% \resizebox{\columnwidth}{!}{
\begin{tabular}{|c|c|c|c|c|c|}
\hline
\textbf{Sample Number}              & \textbf{V profiles} & \textbf{Race line}   & \textbf{Tracks}     & \textbf{Kernels functions} & \textbf{Kernel combinations} \\ [6pt]\hline
\multirow{2}{*}{Full size} &            &             & Shanghai   & RBF               & $ k_{RBF}+(k_{PER}, k_{LIN}) $     \\ [6pt]\cline{5-6} 
                           & Capped     & Center line &            & PERIODIC          & $ k_{RQ}+(k_{PER}, k_{LIN}) $       \\ [6pt]\cline{1-1} \cline{4-6} 
\multirow{2}{*}{Half size} &            &             & Sepang     & RQ                & $ k_{MAT}+(k_{PER}, k_{LIN}) $      \\ [6pt]\cline{2-3} \cline{5-6} 
                           &            &             &            & LINEAR            & $ k_{RBF} \times k_{RQ} $             \\ [6pt]\cline{1-1} \cline{4-6} 
\multirow{2}{*}{1/3 size}  & Non-capped & Race line   & Yas Marina & MATERN            & $ k_{MAT} \times (k_{PER}, k_{RBF}, k_{LIN}) $ \\ [6pt]\cline{6-6} 
                           &            &             &            &                   &  $ k_{RQ} \times (k_{LIN}, k_{MAT}) $    \\ [6pt]\hline
\end{tabular}
% }
\caption{In-depth analysis of GP model schemes.}
\label{tab:analysis}
\end{table}

\subsection{Racing scenarios}
The choices of race line and velocity profiles also influence GP accuracy. Therefore, in the evaluation, we have composed four racing scenarios containing the combinations of these factors.

\subsubsection{RA\_NON-CAP}
\begin{figure}
  \begin{subfigure}[b]{0.48\columnwidth}
    \includegraphics[width=\linewidth]{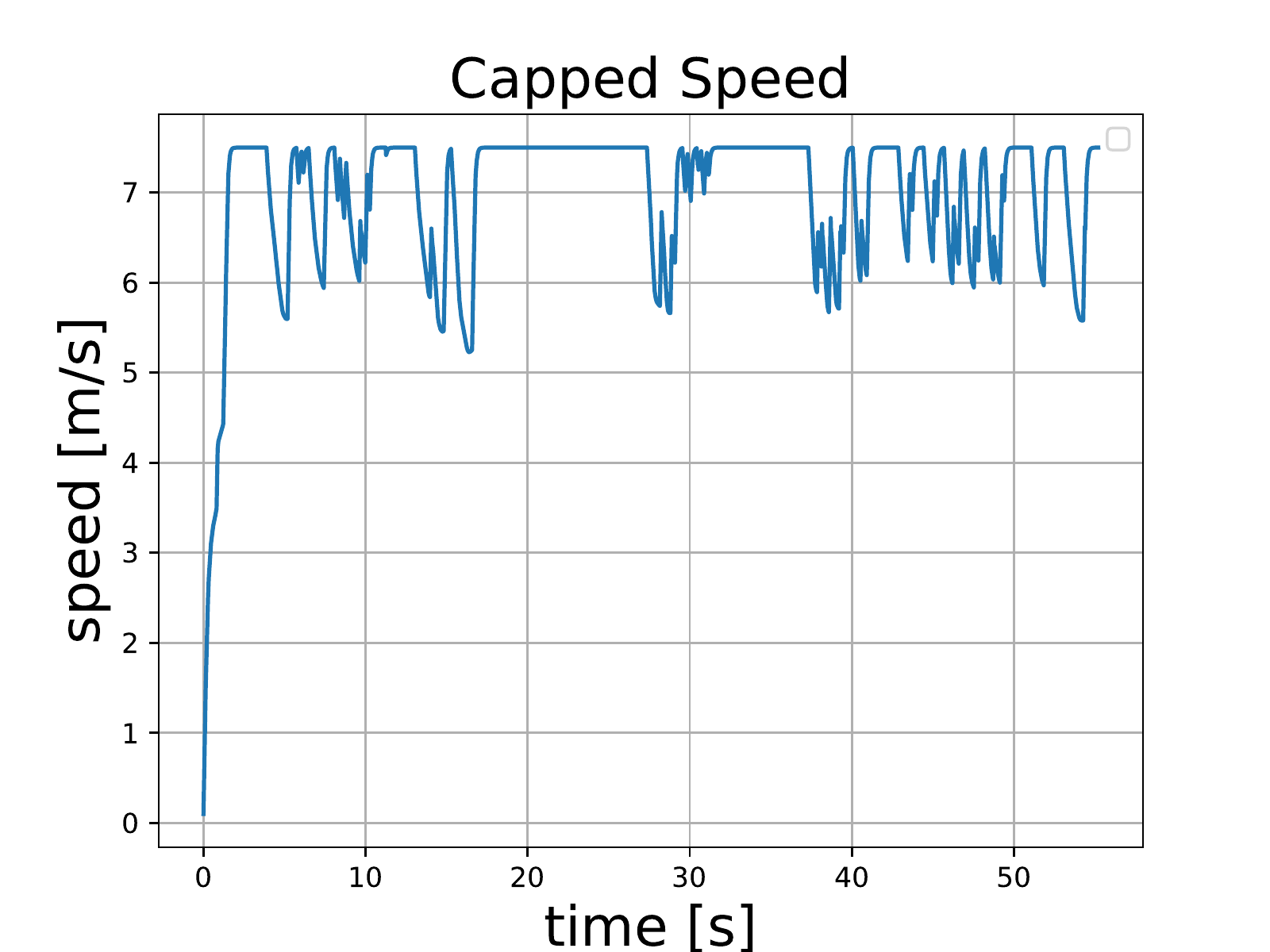}
    \caption{}
    \label{fig:cap}
  \end{subfigure}
  \hfill %%
  \begin{subfigure}[b]{0.48\columnwidth}
    \includegraphics[width=\linewidth]{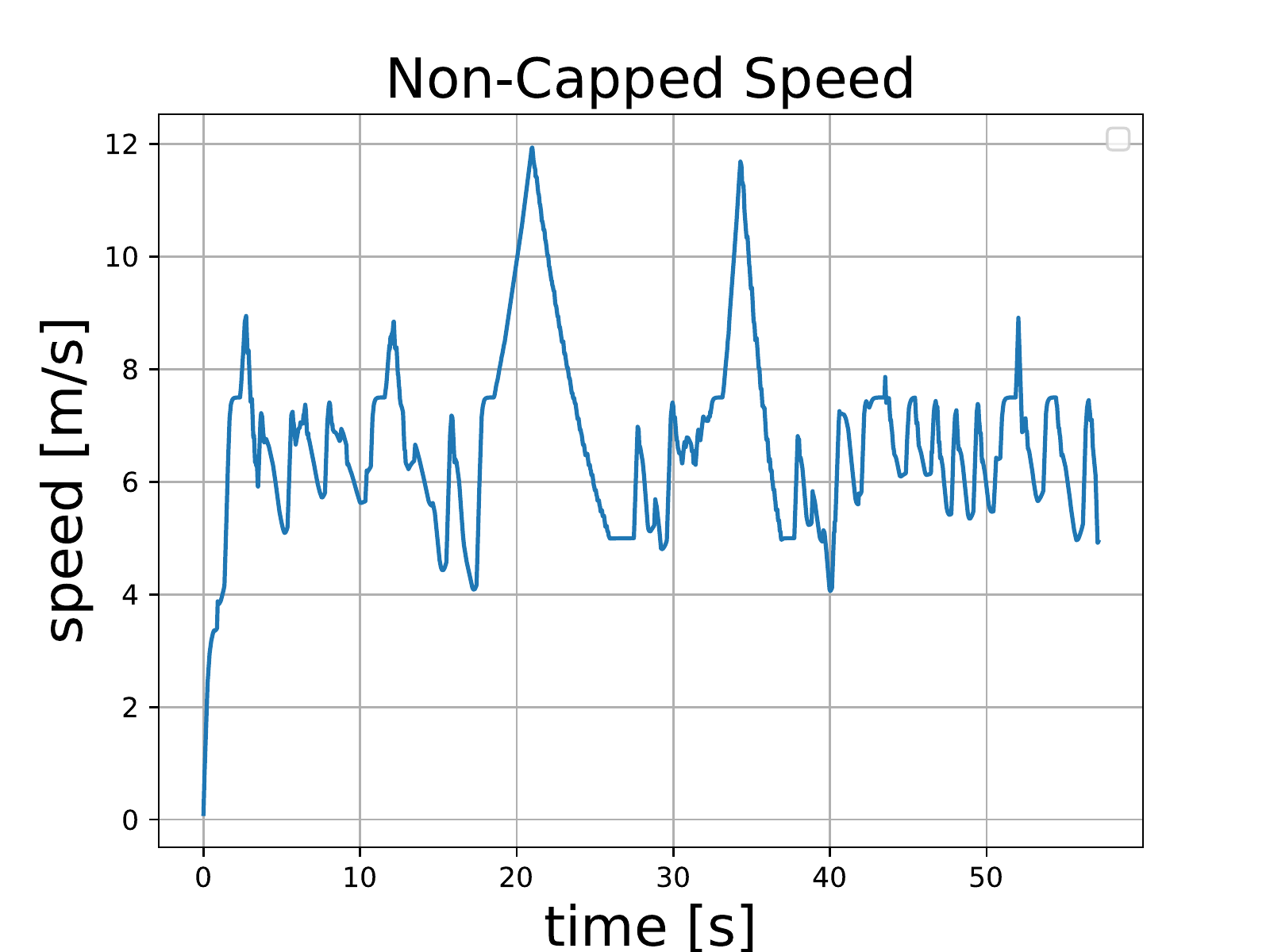}
    \caption{}
    \label{fig:noncap}
  \end{subfigure}
  \caption{Velocity profiles: (a) Capped by a fixed reference speed (e.g. $7.5 [m/s]$); (b) Non-capped speed, dynamic reference speed profile, has linear acceleration and deceleration when driving on straight lines.}
\end{figure}

The vehicle is driving on track to follow an optimal raceline, plus no fixed speed reference to limit the outputs of the pure pursuit controller. In other words, when driving through a long straight track sector, the vehicle will be available to perform a linear speed change of acceleration or deceleration. See in Fig.~\ref{fig:noncap}
\subsubsection{CE\_NON-CAP}
% differentiates with RA\_NON-CAP only in raceline choice. In this scenario, 
The vehicle is driving to follow the center line of each track. The purpose is to compare the GP performances between optimal raceline and simple center line to investigate the importance of calculating raceline in GP model training.
\subsubsection{RA\_CAP}
The vehicle is driving along the optimal raceline as mentioned before but with a fixed reference speed. This will limit the vehicle's top speed in a long straight trajectory. Sometimes a predefined speed profile is not easily accessible with each given track, and it can be non-trivial to create. We design this scenario to show if the GP model performance will decrease when the vehicle is only driving under a fixed speed reference.
\subsubsection{CE\_CAP} The vehicle is driving on the center line with a fixed speed reference. We compare this scenario with both RA\_CAP and CE\_NON-CAP. Thus, we can comprehensively analyze the importance of both speed and raceline profiles in the GP model training process.\par

% To evaluate which scenario generalize the best over the others, we design the Workflow as shown in Fig.~\ref{fig:pipe2}. 

\subsection{Sample sizes in Gaussian Process}
Since GP regression is a non-parametric method, it needs to consider the whole training data each time when making a prediction. This characteristic brings the most prominent weakness of GP: it is computationally expensive. Because of the inversion and determinant of the $n × n$ kernel matrix $K(X,X)$, GP suffers from a cubic time complexity $\mathcal{O}(n^3)$, which limits the scalability of GP. With the inspiration of \cite{liu2020gaussian}, we evaluate the benefits of two types of scalable GPs for vehicle modeling:
(1) Global approximations
approximate the kernel function through global distillation. Specifically, we divide the size of training sample ($n$) into three categories: (i) full size, (ii) half size, and (iii) one-third of full size.
(2) Local approximations
follow the idea of divide-and-conquer to focus on the local subsets of training data. We divide racetrack into three sectors based on their characteristics and approximate the kernel function through each sector. We test our sector-based GP on the whole track dataset.

\section{Experiment Setup}
\subsection{Simulation setup}

\subsubsection{F1TENTH Gym}

\begin{wrapfigure}[22]{r}{0.65\textwidth}
    \centering
    \includegraphics[width=0.65\textwidth]{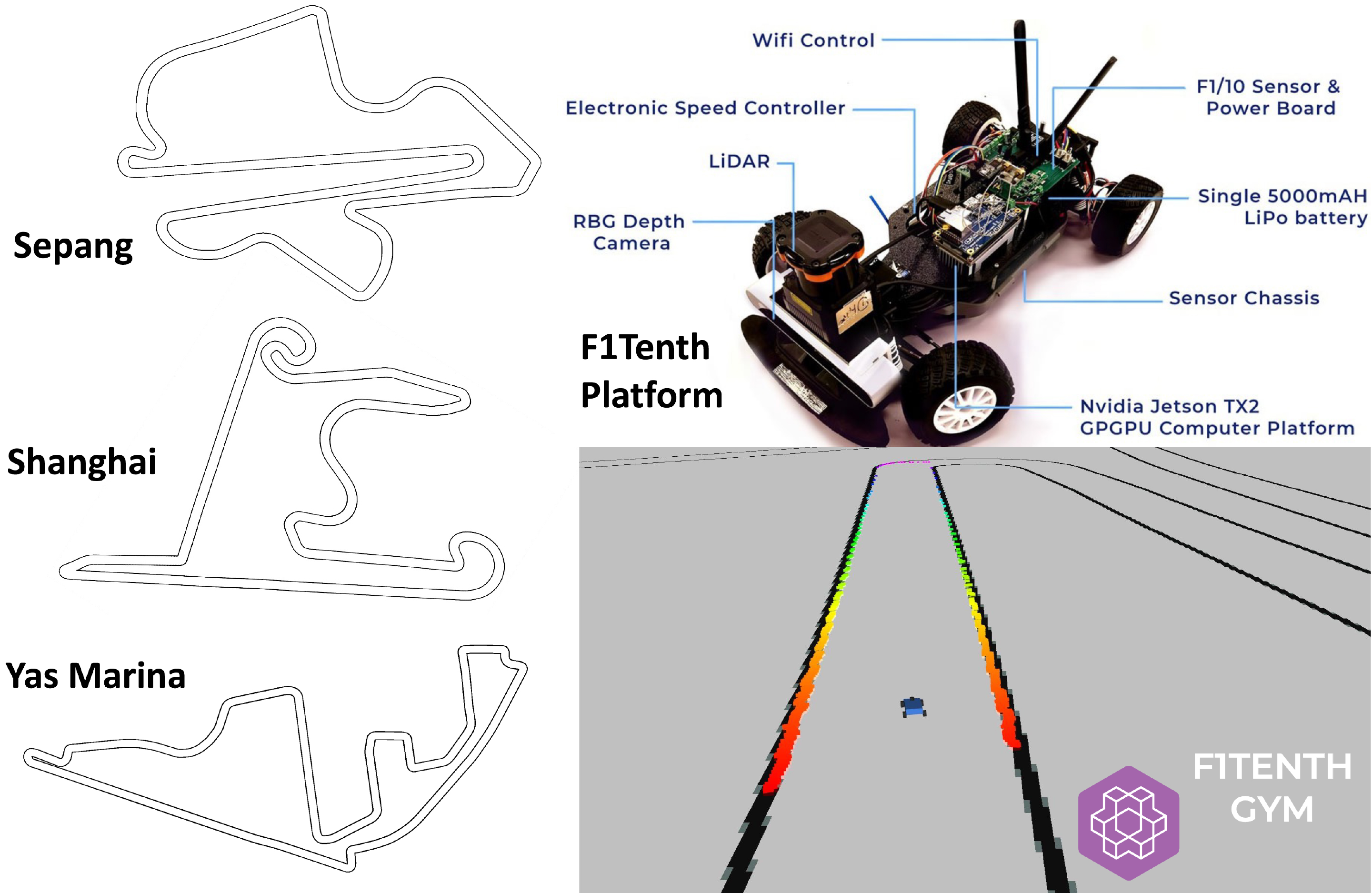}
    \caption{We use the ROS2 based F1TENTH Gym simulator (based on the F1Tenth Platform) in our work on three Formula One Tracks: Sepang International Circuit; Shanghai International Circuit; and Yas Marina Circuit}
    \label{fig:f1tenth_sim}
\end{wrapfigure}

 As shwon in Fig.~\ref{fig:f1tenth_sim}, is created for research that needs an asynchronous, realistic vehicle simulation with multiple vehicle instances in the same environment~\cite{okelly2020f1tenth}. This simulator provides a lightweight, 2D-simulation with an openAI Gym interface. We use ROS2 based F1TENTH Gym as the simulator platform, and all of the simulations are conducted in Unbuntu $20.04.3$ LTS operating system, with an Intel Core i7-10700K CPU.
% The gym environment has been used as the backend for the F1TENTH virtual racing online competition at IROS 2020, as well as used as the simulation engine for the FormulaZero project and TunerCar project.

% \begin{wrapfigure}[16]{r}{0.45\textwidth}
%     \centering
%     \includegraphics[width=0.45\textwidth]{LaTeX/Images/dyn_model_new.pdf}
%     \caption{Single-track dynamic vehicle model. Reference point: C.O.G.}
%     \label{fig:dyn}
% \end{wrapfigure}
% are adopted from F1TENTH racetracks git repo which is based on the f1 racetrack database which includes the racetrack information of 20 real racetracks all around the world. Those data has been downscaled 1:10 to be used in the F1TENTH gym environment. 
In addition, unlike previous work, where researchers made up the racetracks. This paper conduct the experiments on three realistic Formula One racetracks, as shown in Fig.~\ref{fig:f1tenth_sim}, which are: (i) Sepang International Circuit; (ii) Shanghai International Circuit; (iii) Yas Marina Circuit.

% \begin{table}
% \renewcommand{\arraystretch}{1.8}
% \centering
% \resizebox{\columnwidth}{!}{
% \begin{tabular}{|c|c|c|c|}
% \hline
% \textbf{Operating System} & \textbf{CPU}                       & \textbf{RAM} & \textbf{*GPU}           \\ \hline
% Ubuntu 20.04.3 LTS        & Intel Core i7-10700K CPU @ 3.80GHz & 16GB         & NVIDIA GeForce RTX 3080 \\ \hline
% \end{tabular}}
% \caption{Hardware specifications for experiments.}
% \label{tab:Hardware}
% \end{table}
\subsection{Data collection}
We treat the F1tenth Gym environment as the ground truth vehicle dynamics. The model used in the Gym environment is a single-track model from \cite{althoff2017commonroad}.
Thus, we use the collected measurements from Gym environment to address the model mismatch.

The data set has two components, see Table~\ref{equations}: (i) Vehicle states: $[x,y,v,\omega,\beta]$; (ii) Control inputs: $[a_{long}, \delta_v]$.
We collect data by implementing a pure pursuit controller~\cite{coulter1992implementation} in F1TENTH gym environment. For each racetrack, we compute the optimal raceline~\cite{christ2021time} and then track it using the pure pursuit controller. We collect data at a frequency of 60Hz in the form of state-action-state pairs. Dynamic data is denoted by $\mathcal{D}_{dyn} = \{x_k, u_k, x_{k+1}\}, \forall k \in \{0,1,..T-1\}$, where $T$ is the number of data sample size.

\subsection{Training process}
To address the data mismatch between dynamic and E-Kin model data, we use the collected $\mathcal{D}_{dyn}$. Since the parameters of the E-Kin model $f_{kin}$ are known, we generate a new dataset $\mathcal{D}_{kin}$ that captures its response when excited with the same inputs starting from the same initialization. $D_{kin}=\{x_t, u_t, f_{kin}(x_t, u_t)\}, \forall k \in \{0,1,..T-1\}$, where $x_k, u_k$ are states and control inputs data from $D_{dyn}$. Therefore, the training set and model mismatch can be defined as $\mathcal{D} := \mathcal{D}_{dyn} \bigoplus \mathcal{D}_{kin}$, and $e(x_k, u_k) = x_{k+1} - f_{kin}(x_k, u_k)$. The differences between dynamic model and E-Kin model are only in the states $\omega$ and $\beta$, which means error $e$ is of the form $[0,0,0,0,0,*,*]$, where $*$ is nonzero terms. For each state with nonzero error, we train a GP model of the form

\begin{equation}
\begin{aligned}
    e_j := \mathcal{GP}(\omega, \beta, a, \triangle \delta), j \in \{6,7\}
\end{aligned}
\end{equation}

where $j$ corresponds to the model mismatch in states $\omega$ and $\beta$, respectively. Specifically, $e_6 \sim \mathcal{N}(\mu_{\omega}, \sigma_{\omega})$ , and $e_7 \sim \mathcal{N}(\mu_{\beta}, \sigma_{\beta})$

\section{Results}
We organize the evaluation processes of trained GP models in four categories: (i) Evaluation of GP models with different sample sizes, (ii) Evaluation on full track data with different scenarios settings, (iii) Evaluation of GP trained by sector track data, (iv) Evaluation of generalizability of GP model. 
% We present the results of single racetrack in the main paper, the complete results of evaluation processes can be found in the technical appendix.

\subsection{Evaluation metrics}
To validate the performances of GP models, we use: 
\begin{enumerate}
    \item Model training time, $\mathcal{T}$;
    \item Root mean square error, $RMSE$;
    \item Coefficient of determination, $R^2$;
\end{enumerate}

\begin{equation}
\label{eq:R}
    R^2(y, y^*)=1-\frac{\sum_{i=1}^{n}(y_i-y_i^*)^2}{\sum_{i=1}^{n}(y_i-\bar{y})^2}
\end{equation}
where 1 is used to compare the time cost of GP models among different sample sizes, 2 and 3 are used to measure the performance of each GP model in model mismatch estimate. 

\subsection{Result1: Effect of kernel composing}
\begin{table}[h]
\renewcommand{\arraystretch}{1.2}
\centering
\scriptsize
% \resizebox{\columnwidth}{!}{
\begin{tabular}{|c|c|c|c|c|c|}
\hline
\textbf{Tracks}                      & \textbf{Size} & \textbf{RA\_NON-CAP} & \textbf{CE\_NON-CAP} & \textbf{RA\_CAP} & \textbf{CE\_CAP}  \\ [6pt]\hline
\multirow{3}{*}{\textbf{Shanghai}}   & Full          & RQ + LINEAR          & RQ + LINEAR          & MATERN + LINEAR  & RQ + LINEAR       \\ [6pt]\cline{2-6} 
                                     & 1/2           & RQ + LINEAR          & RQ $\times$ LINEAR          & RBF$\times$RQ         & RQ$\times$LINEAR       \\ [6pt]\cline{2-6} 
                                     & 1/3           & RQ + LINEAR          & RQ + LINEAR          & RQ + LINEAR      & RBF$\times$RQ          \\ [6pt]\hline
\multirow{3}{*}{\textbf{Sepang}}     & Full          & RQ + LINEAR          & RQ + LINEAR          & RQ + LINEAR      & MATERN + PERIODIC \\ [6pt]\cline{2-6} 
                                     & 1/2           & RQ + LINEAR          & RQ + LINEAR          & RQ + LINEAR      & RQ$\times$LINEAR       \\ [6pt]\cline{2-6} 
                                     & 1/3           & RQ + LINEAR          & RQ + LINEAR          & RQ + LINEAR      & RBF$\times$RQ          \\ [6pt]\hline
\multirow{3}{*}{\textbf{Yas Marina}} & Full          & MATERN + LINEAR      & MATERN + PERIODIC    & RBF$\times$RQ         & MATERN + PERIODIC \\ [6pt]\cline{2-6} 
                                     & 1/2           & RQ + LINEAR          & RQ + LINEAR          & RQ + LINEAR      & MATERN + PERIODIC \\ [6pt]\cline{2-6} 
                                     & 1/3           & RQ + LINEAR          & RQ + LINEAR          & RQ + LINEAR      & RBF$\times$RQ          \\ [6pt]\hline
\end{tabular}
% }
\caption{Best kernel combination of each racing scenario on each racetrack. All kernels have been trained and tested in GPs with different sample sizes.}
\label{tab:kernelcombo}
\end{table}

Kernel combination is a common solution for GP models that have different inputs, e.g. acceleration and steering velocity, in this paper. 
% Fig.~\ref{fig:pipe1} depicts the workflow of identifying the most versatile kernel function combination for GP models in autonomous race car. 
Here we present the full results of the best kernel combinations for vehicle modeling each scenario in different racetracks in Table~\ref{tab:kernelcombo}.
% The results show that RQ $+$ LINEAR is the most suitable kernel combination across different settings.
With all settings being the same, adding RQ and linear kernel functions produces the best GP model in most scenarios for GP trained on different sample sizes. We thus recommend: $RQ + LINEAR$ kernel is the overall most versatile combination for GP training for autonomous racing. \par
We also notice that the addition operation performs better than the multiplication operation. The kernels that multiply together tend to be difficult to converge in prediction as well as more time-consuming compared to adding kernels together. This may be because of the characteristics of these two operations \cite{duvenaud2014automatic}: (i) Adding kernels can be thought of as an OR operation, i.e., if you add together two kernels, then the resulting kernel will have a high value if either of the two base kernels has a high value. (ii) Multiplying two kernels can be considered an AND operation. That is, if you multiply together two kernels, then the resulting kernel will have a high value only if both of the two base kernels have a high value.

\subsection{Result 2: Effect of sample size on GP}

\begin{table}
\renewcommand{\arraystretch}{1.2}
\centering
\scriptsize
% \resizebox{\columnwidth}{!}{
\begin{tabular}{|c|c|c|c|c|c|}
\hline
\backslashbox{\textbf{Tracks}}{\textbf{Scenarios}}                      &          & \textbf{RA\_NON-CAP} & \textbf{CE\_NON-CAP} & \textbf{RA\_CAP} & \textbf{CE\_CAP} \\ \hline
\multirow{2}{*}{\textbf{Shanghai}}   & training & 2993        & 3042        & 2563    & 2780    \\ \cline{2-6} 
                            & testing  & 3026        & 3079        & 3234    & 3345    \\ \hline
\multirow{2}{*}{\textbf{Sepang}}     & training & 2838        & 3020        & 2597    & 2688    \\ \cline{2-6} 
                            & testing  & 2738        & 2765        & 3083    & 3204    \\ \hline
\multirow{2}{*}{\textbf{Yas Marina}} & training & 2412        & 2543        & 2563    & 2468    \\\cline{2-6} 
                            & testing  & 2632        & 2662        & 2524    & 2600    \\ \hline
\end{tabular}
% }
\caption{Training and testing data sample size of each track under different racing scenarios}
\label{tab:datasize}
\end{table}

We evaluate the performance of GPs with different sample sizes in training time cost and prediction accuracy aspects. 

The details of sample sizes for different simulation settings have been shown in Table~\ref{tab:datasize}. We evaluate the effect of global approximations of GP on training time and prediction accuracy. According to Table~\ref{tab:gptimes}, half-sized downsampling of the data can save around 70\% training time, and one-third of full size can reduce up to 93\% of the full-sized data training time. 
\begin{table}[h]
\renewcommand{\arraystretch}{1.2}
\centering
\scriptsize
% \resizebox{\columnwidth}{!}{
\begin{tabular}{|c|c|c|c|c|c|}
\hline
\textbf{Tracks}                      & \textbf{Size} & \textbf{RA\_NON-CAP} & \textbf{CE\_NON-CAP} & \textbf{RA\_CAP} & \textbf{CE\_CAP} \\ \hline
\multirow{3}{*}{\textbf{Shanghai}}   & Full          & 256.7                & 313.88               & 150.16           & 69.28            \\ \cline{2-6} 
                                     & 1/2           & 65.81                & 58.35                & 28.68            & 32.98            \\ \cline{2-6} 
                                     & 1/3           & 34.31                & 33.12                & 13.48            & 14.9             \\ \hline
\multirow{3}{*}{\textbf{Sepang}}     & Full          & 314.57               & 239.97               & 158.59           & 114.99           \\ \cline{2-6} 
                                     & 1/2           & 58.74                & 38.55                & 30.46            & 32.98            \\ \cline{2-6} 
                                     & 1/3           & 26.19                & 20.85                & 14.02            & 14.9             \\ \hline
\multirow{3}{*}{\textbf{Yas Marina}} & Full          & 201.37               & 245.2                & 121.21           & 75.21            \\ \cline{2-6} 
                                     & 1/2           & 38.66                & 47.49                & 31.5             & 18.4             \\ \cline{2-6} 
                                     & 1/3           & 17.48                & 18.63                & 8.26             & 12.05            \\ \hline
\end{tabular}
% }
\caption{Average training time of GPs with different sample sizes. Training time (in seconds) has been calculated in different scenarios across different racetracks.}
\label{tab:gptimes}
\end{table}

Table~\ref{tab:gpev} presents the prediction performance, which has shown that compared with full-sized GPs, training GPs with $50\%$ less data only decreases model accuracy by $2\%$. Moreover, training with $66\%$ less data only decreases the accuracy by $4\%$.  \par

\begin{table}[]
\renewcommand{\arraystretch}{1.2}
\centering
\scriptsize
% \resizebox{\columnwidth}{!}{
\begin{tabular}{|c|c|c|c|c|c|}
\hline
\textbf{Tracks}                      & \textbf{Size} & \textbf{RA\_NON-CAP} & \textbf{CE\_NON-CAP} & \textbf{RA\_CAP} & \textbf{CE\_CAP} \\ \hline
\multirow{3}{*}{Shanghai}   & Full                   & 0.979                         & 0.989                         & 0.975                     & 0.973                     \\ \cline{2-6} 
                            & 1/2                    & 0.965                         & 0.976                         & 0.965                     & 0.968                     \\ \cline{2-6} 
                            & 1/3                    & 0.953                         & 0.954                         & 0.944                     & 0.951                     \\ \hline
\multirow{3}{*}{Sepang}     & Full                   & 0.978                         & 0.989                         & 0.977                     & 0.981                     \\ \cline{2-6} 
                            & 1/2                    & 0.977                         & 0.981                         & 0.975                     & 0.968                     \\ \cline{2-6} 
                            & 1/3                    & 0.972                         & 0.95                          & 0.963                     & 0.951                     \\ \hline
\multirow{3}{*}{Yas Marina} & Full                   & 0.971                         & 0.962                         & 0.951                     & 0.952                     \\ \cline{2-6} 
                            & 1/2                    & 0.958                         & 0.938                         & 0.95                      & 0.947                     \\ \cline{2-6} 
                            & 1/3                    & 0.941                         & 0.926                         & 0.932                     & 0.921                     \\ \hline
\end{tabular}
% }
\caption{Performance downgrade of global approximations of GP models. The value is calculated by mean values of explaining variances under different scenarios of each racetrack.}
\label{tab:gpev}
\end{table}

The global approximations of GP models can significantly reduce the time cost of training processes while remain good performance in the meantime. Therefore, if time is limited, we recommend using half or less data to train GP models. 

\subsection{Result 3: Effect of racing scenarios on GP}

\begin{figure}
    \centering
    \includegraphics[width=0.7\columnwidth]{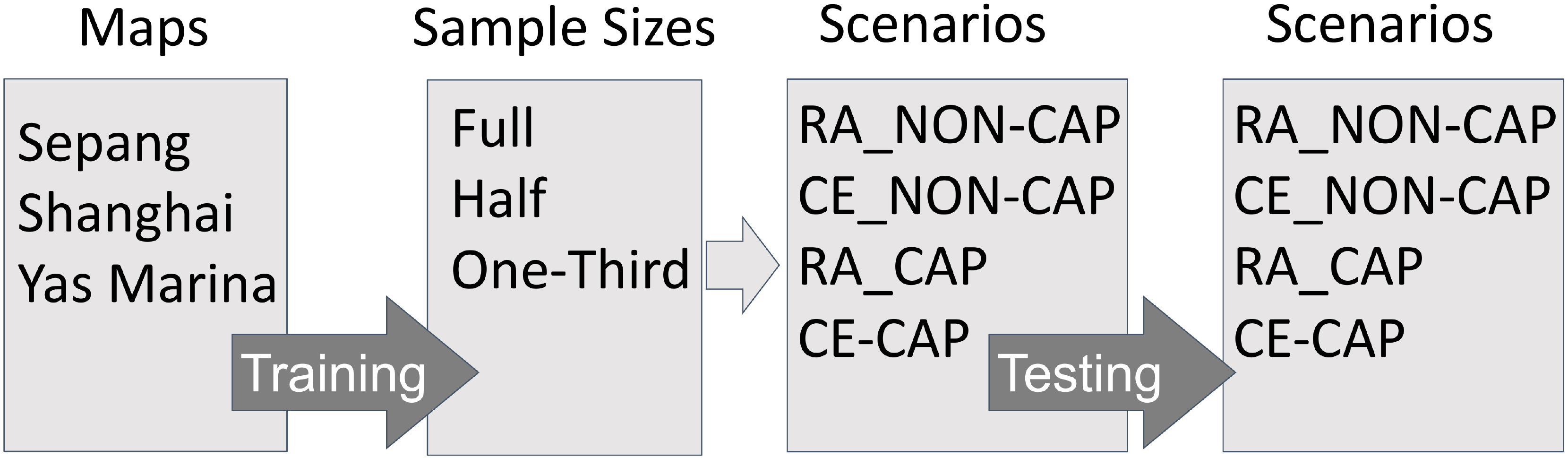}
\caption{Evaluation of the best racing scenarios}
\label{fig:pipe2}
\end{figure}

To evaluate the generalizability of different scenarios, we design the workflow as shown in Fig.~\ref{fig:pipe2}. The training process is as same as before, but we compose a larger test set consisting of full-sized data sample and all the racing scenarios. Besides, we use the kernel combination yields the best performance of each scenario for testing.

\begin{table}[h]
\renewcommand{\arraystretch}{1.2}
\scriptsize
\centering
% \resizebox{\columnwidth}{!}{
\begin{tabular}{|c|c|c|c|c|c|}
\hline
\textbf{Tracks}                      & \textbf{Size} & \textbf{RA\_NON-CAP} & \textbf{CE\_NON-CAP} & \textbf{RA\_CAP} & \textbf{CE\_CAP} \\ \hline
\multirow{3}{*}{\textbf{Shanghai}}   & Full          & 0.988                & 0.989                & 0.982            & 0.983            \\ \cline{2-6} 
                                     & 1/2           & 0.988                & 0.987                & 0.981            & 0.983            \\ \cline{2-6} 
                                     & 1/3           & 0.984                & 0.985                & 0.969            & 0.972            \\ \hline
\multirow{3}{*}{\textbf{Sepang}}     & Full          & 0.991                & 0.991                & 0.963            & 0.968            \\ \cline{2-6} 
                                     & 1/2           & 0.99                 & 0.989                & 0.955            & 0.960            \\ \cline{2-6} 
                                     & 1/3           & 0.984                & 0.981                & 0.950            & 0.951            \\ \hline
\multirow{3}{*}{\textbf{Yas Marina}} & Full          & 0.963                & 0.965                & 0.961            & 0.953            \\ \cline{2-6} 
                                     & 1/2           & 0.954                & 0.956                & 0.945            & 0.939            \\ \cline{2-6} 
                                     & 1/3           & 0.912                & 0.923                & 0.917            & 0.867            \\ \hline
\end{tabular}
% }
\caption{Performance of the best GP models of each scenario over the whole test set. }
\label{tab:scepfm}
\end{table}

As shown in Table~\ref{tab:scepfm}, a dynamic speed reference is preferred since they always generalize better than the ones with fixed speed, and fixed speed profiles tend to cause high uncertainty. In addition, results also show that there are no significant benefits between the raceline and centerline. The centerline of a racetrack is easier to obtain, while the raceline, on the other hand, requires extra calculations of either minimal lap time or the shortest path. Therefore, using a dynamic speed profile with a centerline track file is sufficient for training a decent GP model.

\subsection{Result 4: In-depth sectors analysis}
We provide an in-depth analysis of local approximations, where we select three sectors of each racetrack and trained the GPs based on them.\par
\begin{table}[h]
\renewcommand{\arraystretch}{1.2}
\centering
\scriptsize
% \resizebox{\columnwidth}{!}{
\begin{tabular}{|c|c|c|c|c|c|}
\hline
\textbf{Tracks}                      & \textbf{Sectors} & \textbf{RA\_NON-CAP} & \textbf{CE\_NON-CAP} & \textbf{RA\_CAP} & \textbf{CE\_CAP} \\ \hline
\multirow{3}{*}{Shanghai}   & 1                         & 0.888                         & 0.973                         & -1.57e+09        & -4.106e+20       \\ \cline{2-6} 
                            & 2                         & 0.979                         & 0.98                          & -5.01e+10        & 0.566*           \\ \cline{2-6} 
                            & 3                         & 0.986                         & 0.987                         & 0.984            & 0.989            \\ \hline
\multirow{3}{*}{Sepang}     & 1                         & 0.975                         & 0.904                         & -8.94            & 0.566*           \\ \cline{2-6} 
                            & 2                         & 0.989                         & 0.976                         & -5.316e+12       & 0.983*           \\ \cline{2-6} 
                            & 3                         & 0.991                         & 0.99                          & 0.984            & 0.983            \\ \hline
\multirow{3}{*}{Yas Marina} & 1                         & 0.903                         & 0.874                         & 0.955*           & 0.92*            \\ \cline{2-6} 
                            & 2                         & 0.918                         & 0.899                         & 0.865*           & 0.78*            \\ \cline{2-6} 
                            & 3                         & 0.959                         & 0.942                         & 0.961            & 0.935            \\ \hline
\end{tabular}
% }
\caption{In-depth sectors analysis. Performance comparison among sectors under different scenarios of each racetrack. ( $*$ denotes models with high uncertainty.)}
\label{tab:sectpfm}
\end{table}

\begin{figure}
% \centering
  \begin{subfigure}[]{0.45\columnwidth}
  \centering
    \includegraphics[width=\columnwidth]{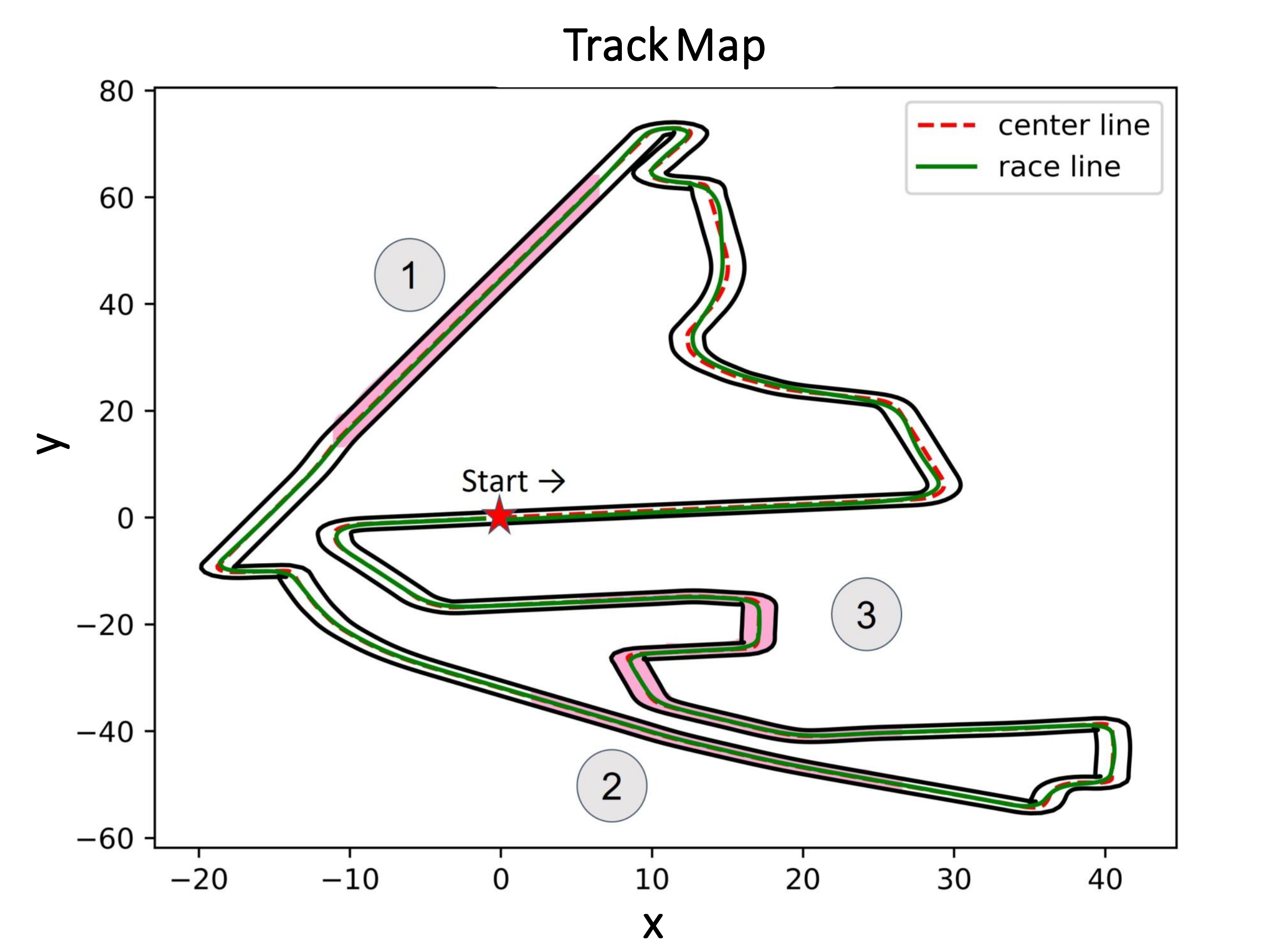}
    \caption{}
   
  \end{subfigure}
  % \newline
%   \hfill %%
  \begin{subfigure}[]{0.45\columnwidth}
    \includegraphics[width=\columnwidth]{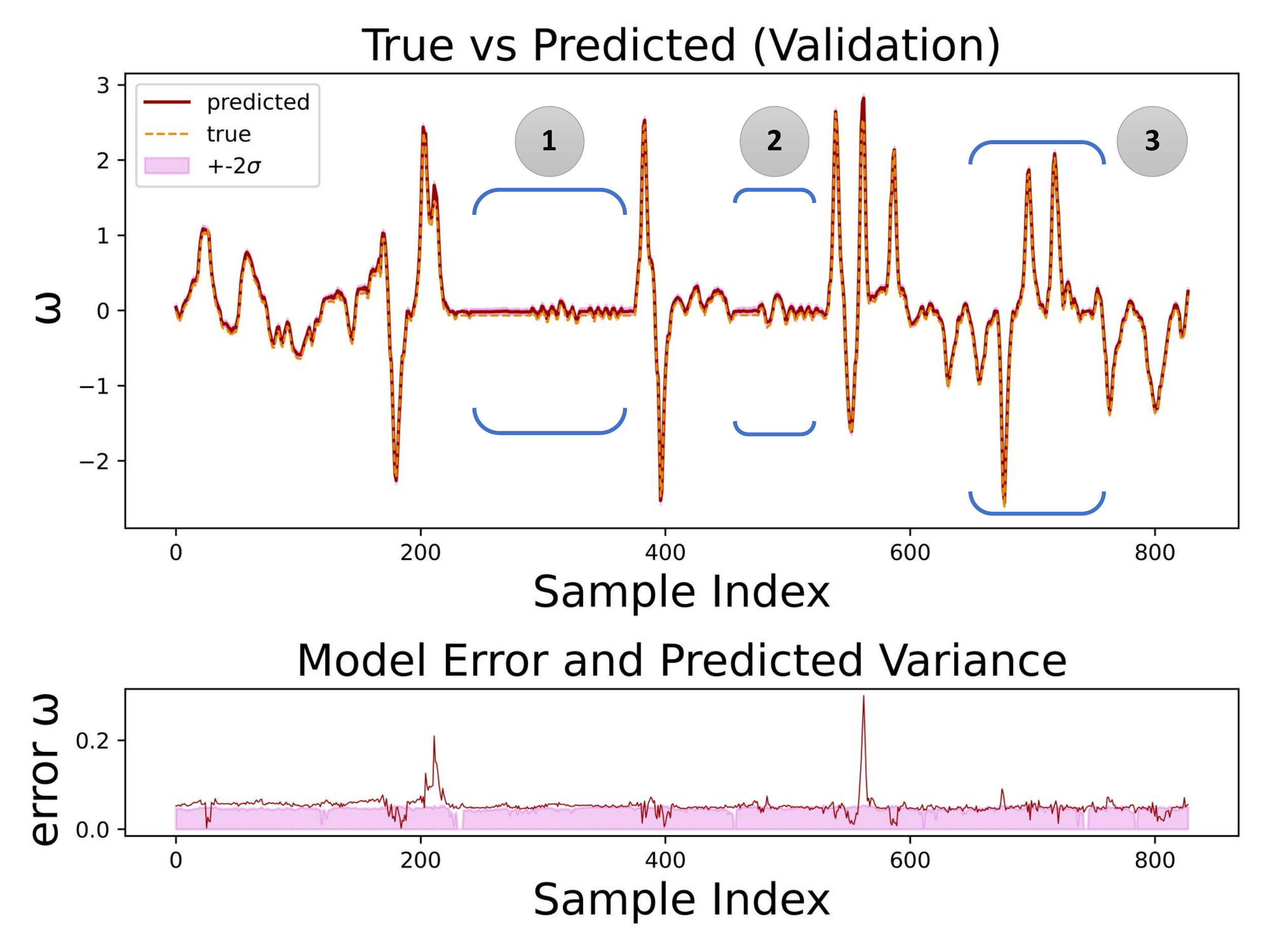}
    \caption{}
    \label{fig:sec2}
  \end{subfigure}
  \caption{Sector analysis: (a)three sectors of racetrack Yas Marina. (b) GP model predictions of corresponding sectors.}
  \label{fig:sec1}
\end{figure}

One of our hypotheses is that a curve trajectory is better suited for training GPs than a straight trajectory where there is little variation in the data, i.e., yaw rate, $\omega$ and body slip angles, $\beta$. Therefore, in addition to the global approximations, we evaluate GPs on local subset of data. Specifically, we focus on three sectors of each racetrack. Fig.~\ref{fig:sec1} shows chosen sectors of Yas Marina Circuit. To introduce the path variation, these sectors are composed of: (i) straight lines, where the error terms, $e_\omega$, $e_\beta$, are almost constant; (ii) straight line with minor steering, causes a little variation in the error terms; (iii) curvy paths and cornering, which fluctuate error terms, $e_\omega$ and $e_\beta$. Although the training processes are conducted on local subsets, the test sets are full-sized data for each experiment. The kernel combination has been selected to be the best kernel for corresponding settings. 
Based on Table~\ref{tab:sectpfm}, the local approximations-based GP models yield accurate results as long as we select the correct representations. Specifically, GP models trained by curvy trajectories always have better prediction performances comparing to straight sectors. Besides, sector-trained GPs with CAP speed profile on straight trajectories tend to cause high model uncertainties. This also explains that why previous work of GP regression in vehicle dynamics modeling prefers to conduct experiments with artificial racetracks which seldom conclude straight line trajectories.

\subsection{Result 5: Generalizing local approximations}

We next evaluate the generalizability of local approximated GP models by testing them on racetracks other than the one used to train the models. In other words, we test the GP models trained on sectors on new racetracks that they have never seen. As the results shown in Table~\ref{tab:sectpfm} indicates that GP regression models trained on curvier sectors always performs better than less curvy and straight sectors. Therefore, we select the third sector from each racetrack with combination of raceline and dynamic speed profile in the evaluation processes. 

\begin{table}[]
\renewcommand{\arraystretch}{1.22}
\scriptsize
\centering
% \resizebox{\columnwidth}{!}{
\begin{tabular}{|c|c|c|c|c|c|}
\hline
\textbf{Tracks}                      & \textbf{Testing tracks} & \textbf{RA\_NON-CAP} & \textbf{CE\_NON-CAP} & \textbf{RA\_CAP} & \textbf{CE\_CAP} \\ \hline
\multirow{2}{*}{\textbf{Shanghai}}   & Sepang                  & 0.989                & 0.99                 & 0.989            & 0.987            \\ \cline{2-6} 
                                     & Yas Marina              & 0.962                & 0.943                & 0.965            & 0.95             \\ \hline
\multirow{2}{*}{\textbf{Sepang}}     & Shanghai                & 0.982                & 0.984                & 0.975            & 0.98             \\ \cline{2-6} 
                                     & Yas Marina              & 0.946                & 0.921                & 0.961            & 0.937            \\ \hline
\multirow{2}{*}{\textbf{Yas Marina}} & Sepang                  & 0.991                & 0.991                & 0.99             & 0.987            \\ \cline{2-6} 
                                     & Shanghai                & 0.987                & 0.989                & 0.981            & 0.986            \\ \hline
\end{tabular}
% }
\caption{Sectors Generalization. In each racetrack, we train GP models on a local subset of the track using raceline + dynamic speed profile and test models on the other racetracks they have never seen.}
\label{tab:sectpgen}
\end{table}

Based on Table~\ref{tab:sectpgen}, GP models trained on sectors generalize well across racetracks they have never seen. Noted we use the best kernel combinations of each racing scenario for this evaluation process. Although the sector-trained GP perform slightly worse, $2\% - 5\%$, than the full-sized models in prediction accuracy, they are much more computing efficient and can save $90\%$ of the training time than full-sized GP models.

\section{Conclusion and Discussion}
% Although GP regression models perform well in model mismatch prediction on error term $e_\omega$, we notice that they struggled on data with little variation. The values of body slip angle, $\beta$, not only have a small range of values, varying between $-0.06$ and $-0.01$, but also have flat regions where the vehicle was driving on straight lines. In this case, the GP models are hard to converge or align with the ground truth. However, GP can still predict the pattern of ground truth, which makes the error between prediction and ground truth a constant value most time.\par

% This evaluation provides another caution on data selection for implementing the GP model. In addition to the size of the data sample, in which a large data set will cause high computing costs, the performance of GP on vehicle modeling also suffers from data in small values.

In this paper, we implement a pure pursuit controller on the 1/10 scale racecar in a gym environment to drive on three real-world racetracks for data collection and then train GP regression models to learn model mismatch between single-track and extended kinematic models. Besides, we present a thorough study of vehicle dynamics modeling for autonomous racing using Gaussian Processes. For modeling evaluation, we have conducted the following experiments: (i) evaluation of kernel composing; (ii) evaluate the effect of global approximations; (iii) evaluate the effect of racing scenarios; (iv) an in-depth sectors analysis of each racetrack; (v) evaluation of the generalizability of each sector; (vi) limitation of GP regression models in small value data.

In the future, we will continue on evaluating the importance of GP models in vehicle dynamic modeling. We plan to implement advanced control algorithms on the corrected model: 
$f_{corr}(x_k,u_k) = f_{kin}(x_k,u_k) + e(x_k,u_k)$.By comparing the control performance between corrected vehicle models and real dynamics models, we will clearly understand the capability of GP regression in vehicle dynamic modeling for autonomous racing.

\newpage
\bibliography{root}  
\end{document}